\documentclass[twoside,11pt]{article}

\usepackage{blindtext}
\usepackage{listings}
\usepackage{xcolor}
\usepackage[preprint]{jmlr2e}
\usepackage{float}
\usepackage{lastpage}
\jmlrheading{23}{2022}{1-\pageref{LastPage}}{1/21; Revised 5/22}{9/22}{21-0000}{Author One and Author Two}

\ShortHeadings{FSEVAL}{FSEVAL}
\firstpageno{1}

\begin{document}

\title{FSEVAL: Feature Selection Evaluation Toolbox and Dashboard}

\author{\name Muhammad Rajabinasab \email rajabinasab@imada.sdu.dk \\
       \addr Department of Mathematics and Computer Science\\
       University of Southern Denmark\\
       Odense, Denmark
       \AND
\name Arthur Zimek \email zimek@imada.sdu.dk \\
       \addr Department of Mathematics and Computer Science\\
       University of Southern Denmark\\
       Odense, Denmark}

%\editor{My editor}

\maketitle

\begin{abstract}%   <- trailing '%' for backward compatibility of .sty file
Feature selection is a fundamental machine learning and data mining task, involved with discriminating redundant features from informative ones. It is an attempt to address the curse of dimensionality by removing the redundant features, while unlike dimensionality reduction methods, preserving explainability. Feature selection is conducted in both supervised and unsupervised settings, with different evaluation metrics employed to determine which feature selection algorithm is the best. In this paper, we propose FSEVAL, a feature selection evaluation toolbox accompanied with a visualization dashboard, with the goal to make it easy to comprehensively evaluate feature selection algorithms. FSEVAL aims to provide a standardized, unified, evaluation and visualization toolbox to help the researchers working in the field, conduct extensive and comprehensive evaluation of feature selection algorithms with ease.
\end{abstract}


\section{Introduction}
Feature selection is a fundamental machine learning and data mining task. The objective of feature selection is to identify the most relevant features for representing $y$ while eliminating redundant or irrelevant dimensions, thereby improving model generalization and reducing computational costs by overcoming the \emph{curse of dimensionality} \citep{guyon2003introduction,li2017feature}.  Compared to dimensionality reduction methods, feature selection offers a huge advantage; keeping the explainability properties of the original feature space despite reducing the dimensionality. This is why, despite the existence of successful methods for dimensionality reduction, such as Principal Component Analysis (PCA) \citep{pearson1901}, Autoencoders \citep{bourlard1988}, t-distributed Stochastic Neighbor Embedding (t-SNE) \citep{vandermaaten2008}, and Uniform Manifold Approximation and Projection (UMAP) \citep{mcinnes2018}, feature selection methods are still important, and in many cases preferred, to tackle the \emph{curse of dimensionality} \citep{guyon2003introduction,dy2003feature, dai2026online,zou2026feature,jain2026enfestdroid}.

The evaluation of feature selection is most often conducted by a downstream task. In supervised learning, it is often measured using classification accuracy (ACC) and Area Under the Curve (AUC), reflecting the model's predictive power using the selected subset. For unsupervised scenarios, algorithms are often evaluated using $k$-Means clustering \citep{lloyd1982}, using metrics such as Clustering Accuracy (CLSACC) and Normalized Mutual Information (NMI), which assess the alignment between the discovered clusters and the underlying data structure \citep{rajabinasab2024fsdem, mostert2021feature}. Recent advancements have introduced more comprehensive frameworks to address the limitations of these standard metrics. The Feature Selection Dynamic Evaluation Metric (FSDEM) offers a dynamic assessment by tracking performance across varying subset sizes \citep{rajabinasab2024fsdem}, while the Average Angle Difference (AAD) provides a model-independent evaluation based on the alignment of principal components between the original and reduced feature spaces \citep{rajabinasab2025metrics}. The Baseline Fitness Improvement (BFI) normalizes performance against the full feature set, ensuring that any gain is attributed to the selection quality rather than dataset simplicity \citep{mostert2021feature}. Furthermore, the stability of these selections is essential for ensuring robustness; foundational indices by \citet{kuncheva2007stability} and the more rigorous consistency measures by \citet{nogueira2017stability} quantify how sensitive the selection process is to data perturbations. And more importantly, the stability from an informational perspective is also offered by the FSDEM metric \citep{rajabinasab2024fsdem}.

In this paper, we propose FSEVAL\footnote{Python source code provided at: https://github.com/mrajabinasab/FSEVAL}, a comprehensive toolbox for the extensive evaluation of feature selection algorithms. Programmed purely using Python, FSEVAL provides all the necessary tools to evaluate feature selection algorithms using supervised and unsupervised tasks and metrics, model-independent AAD metric \citep{rajabinasab2025metrics}, calculate FSDEM score \citep{rajabinasab2024fsdem}, 
and conducting comprehensive runtime analysis. FSEVAL comes with a visualization dashboard, which can easily load the evaluation output and offer visualizations for different metrics, runtime analysis, and additionally, valuable insights such as rank analysis and Critical Difference (CD) diagrams based on both the standard rank statistics \citep{demsar2006statistical} and MARS~\citep{rajabinasab2026marsmagnitudeawarerankstatistics}.

\section{FSEVAL Toolbox and Dashboard}

FSEVAL addresses a critical gap in machine learning research: the lack of a standardized, rigorous framework for the evaluation and comparison of the feature selection algorithms. While many feature selection methods exist, evaluating them often involves fragmented scripts that fail to account for stochastic variance or how a method scales as the number of features increases. FSEval provides a unified pipeline that automates the evaluation and comparison, allowing researchers to evaluate their feature selection methods of interest, using their desired datasets, in a robust and comprehensive evaluation setup. The FSEVAL dashboard helps with easy and interactive visualization of the evaluation outputs.

\subsection{FSEVAL Toolbox}

FSEVAL is built in a modular format that treats feature selection as a dynamic process rather than a static result. Instead of conducting a single-point evaluation, which often hides many aspects of the feature selection process, FSEval evaluates performance across a spectrum of feature sizes—often referred to as ``feature ratio grids''. This reveals the ``elbow point'' of an algorithm, identifying the exact moment where removing more features begins to degrade model accuracy, as well as yielding insights into the overall feature selection process, which can be easily reflected by feeding the evaluation output of FSEVAL to FSDEM \citep{rajabinasab2024fsdem}. FSEVAL conducts experiments on the feature selection process by default on the first 10\%, by a step size of 0.5\%, and the entire range, by a step size of 5\%, to test the feature selection performance both in the entire process, and also in extensive and extreme feature selection scenarios.

There are different evaluation metrics built in to the FSEVAL toolbox:
\begin{itemize}
    \item \textbf{Supervised Evaluation:} Assesses the predictive power of a feature subset via cross-validated classification performance, utilizing metrics such as Accuracy (ACC) and the Area Under the ROC Curve (AUC) \citep{li2017feature}.
    
    \item \textbf{Unsupervised Evaluation:} Measures the accordance of the dimensionality-reduced data structure to the external labels. This is quantified through clustering-based metrics, including Clustering Accuracy (CLSACC) and Normalized Mutual Information (NMI) \citep{li2017feature}.
    
    \item \textbf{Model-Agnostic Evaluation:} Quantifies the quality of the feature selection process by calculating AAD \citep{rajabinasab2025metrics}.

    \item \textbf{Stability:} Uses FSDEM \citep{rajabinasab2024fsdem} to measure the informational stability of the feature selection algorithm. This process requires the evaluation results of at least one metric of interest.

    \item \textbf{Custom Evaluation:} FSEVAL supports easy integration of \emph{custom} metrics, which can be easily defined as a function and fed to the evaluation procedure.

\end{itemize}

Besides the evaluation metrics, FSEVAL also facilitates conducting \textbf{Runtime Analysis}, by stress-testing feature selection algorithms using data with different number of instances and features, highlighting to which a method is more sensitive. 

\subsection{FSEVAL Dashboard}
The FSEVAL dashboard is designed to provide easy, comprehensive, and interactive visualization for the evaluation results produced by the FSEVAL toolbox. An already online version of this dashboard is accessible at: \url{https://fseval.imada.sdu.dk}. This dashboard can automatically process the evaluation results from FSEVAL toolbox, and create visualizations and tables with valuable insights. All the datasets used in the experiments behind the dashboard are selected from the scikit-feature repository \citep{li2017feature}.
The primary capabilities provided by the dashboard include:
\begin{itemize}
    \item \textbf{Multi-Metric Performance Profiling:} Generates interactive curves that visualize how evaluation metrics (ACC, AUC, CLSACC, NMI, AAD, and Stability)
    evolve across the feature ratio grid. This enables the identification of the optimal subset size where performance saturates. In addition to that, a table is available for the FSDEM scores \citep{rajabinasab2024fsdem} of both 10\% and 100\% experiments 
    for every metric. Additionally, a rank analysis is automatically conducted based on both the standard rank statistics~\citep{demsar2006statistical} and MARS~\citep{rajabinasab2026marsmagnitudeawarerankstatistics}, and critical difference diagram visualization is provided. All figures and tables are easy to export as \textit{publication-ready} PDF file or latex code.
    
    \item \textbf{Computational Efficiency Mapping:} Visualizes results from the scalability \texttt{timer} module, plotting execution time against increasing feature dimensionality and instance counts. This allows for a direct trade-off analysis between predictive utility and algorithmic complexity, as well as pointing out the sensitivity direction of the algorithms.

    \item \textbf{Additional Features:} Many additional features are also implemented, including  seamless importation of benchmark results, the random baseline~\citep{rajabinasab2026worserandomimportancebaseline} to guide the development and the evaluation of unsupervised feature selection methods, customizing the results by selecting the combination of the available methods and datasets, and a web page to download currently presented raw result files and datasets. 
    
\end{itemize}

\subsection{Related Work}
To the best of our knowledge, no prior work provides a comprehensive benchmark tool for feature selection algorithms. The closest to our work might be featsel \citep{REIS201710}, which provides a mathematically rigorous C++ environment for benchmarking the efficiency of search algorithms and cost functions within a Boolean lattice. The primary focus of featsel remains on the computational optimization of the search trajectory itself. In contrast, FSEval serves a distinct role by providing an easy way to conduct extensive and robust evaluation of feature selection algorithms using a variety of metrics and from different aspects. On the other hand, there is no other tool providing a dashboard to visualize and process feature selection evaluation results into a variety of informative visualizations and tables, with the option to export publication-ready copies.

\section{Conclusion and Future Works}
In this paper we proposed and described FSEVAL, a toolbox and dashboard for conducting extensive evaluation and analyses on feature selection algorithms with ease. We discussed different metrics and experiments built into the toolbox, as well as the flexibility to easily introduce and integrate new metrics with the experimental setup. We also discussed the visualization dashboard which processes the toolbox evaluation results into insightful visualizations, ready to be exported for scientific publications. We also deployed the dashboard online, alongside with the evaluation results of a variety of feature selection algorithms on an extensive selection of datasets.

As for the FSEVAL ecosystem, we aim to keep both the dashboard and the toolbox updated with new metrics, methods, and analyses. We  actively encourage fellow researchers in the field of feature selection to conduct their evaluations with FSEVAL to have a comprehensive and standardized benchmark. We also encourage researches to share their evaluation results with us via \href{mailto:fseval@imada.sdu.dk}{fseval@imada.sdu.dk}. Upon receiving and verifying the results, we will include their algorithm and its results in the online dashboard, further enriching the field by providing real-time comprehensive insights onto the progress of its development.

%\newpage
\acks{This study was funded by Innovation Fund Denmark in the project ``PREPARE: Personalized Risk Estimation and Prevention of Cardiovascular Disease''.}

\newpage

\appendix

\section{Installation and Usage}
\texttt{FSEval} is comprised of a modular Python-based toolbox for automated benchmarking and a companion web-based dashboard for interactive analysis. Together, they provide a systematic pipeline designed to answer which feature selection methods perform best by facilitating the experiments and analyses.

\subsection{FSEval Toolbox}

The core library is a lightweight framework that automates the rigorous evaluation of feature selection and ranking methods. It is designed to provide an easy way for comprehensive and robust evaluation of feature selection algorithms.

\subsubsection{Installation and Requirements}
\texttt{FSEval} requires Python 3 and depends on the standard scientific stack, including \texttt{numpy}, \texttt{pandas}, \texttt{scikit-learn}, \texttt{scipy}, \texttt{pcametric}, \texttt{fsdem}, and \texttt{clustpy}. The library can be installed via the Python Package Index or just simply imported from GitHub.
\begin{verbatim}
pip install sdufseval
\end{verbatim}

\subsubsection{API Architecture}
The benchmarking process is managed by the \texttt{FSEVAL} class. It allows for high-granularity control over the evaluation environment, as detailed in Table~\ref{tab:init_params}.

\begin{table}[ht]
\centering
\caption{FSEval Initialization Parameters}
\label{tab:init_params}
\begin{tabular}{llp{8.5cm}} % 'p' sets a fixed width and enables wrapping
\hline
\textbf{Parameter} & \textbf{Default} & \textbf{Description} \\ \hline
\texttt{output\_dir} &  ``results'' & Destination folder for CSV result persistence. \\
\texttt{cv} & 5 & Number of cross-validation folds (supervised). \\
\texttt{avg\_steps} & 10 & Repetitions for stochastic methods. \\
\texttt{supervised\_iter} & 5 & Classifier runs with distinct random seeds. \\
\texttt{unsupervised\_iter} & 10 & Clustering runs with distinct random seeds. \\
\texttt{eval\_type} & \textit{List} & ["supervised", "unsupervised", "model\_agnostic"]. \\
\texttt{metrics} & \textit{List} & ["CLSACC", "NMI", "ACC", "AUC", "AAD"]. \\
\texttt{stability} & True & Whether to calculate stability based on FSDEM score for all metrics or not. ``True'', ``False'', or a list of metrics to calculate stability for. \\
\texttt{custom\_metrics} & \{\} & User-defined dictionary of metric callables. \\
\texttt{experiments} & \textit{List} & Target feature ratios (e.g., ["10Percent", "100Percent"]). \\
\texttt{save\_all} & False & Save individual stochastic runs separately. \\ \hline
\end{tabular}
\end{table}

\subsubsection{Core Methods and Execution}
The library's functionality is primarily accessed through the \texttt{run} and \texttt{timer}, which are used to run the main experiments and runtime analyses respectively. These functions handle the iteration over datasets and the recording of performance and scalability metrics, as summarized in Table~\ref{tab:core_methods}.

\begin{table}[tb]
\centering
\caption{Core API Methods and Arguments}
\label{tab:core_methods}
\begin{tabular}{lll}
\hline
\textbf{Method} & \textbf{Argument} & \textbf{Description} \\ \hline
\texttt{run()} & \texttt{datasets} & List of dataset names loadable via \texttt{load\_dataset()}. \\
 & \texttt{methods} & List of dicts: \texttt{\{'name', 'func', 'stochastic'\}}. \\
 & \texttt{classifier} & Sklearn estimator (Default: \texttt{RandomForestClassifier}). \\ \hline
\texttt{timer()} & \texttt{methods} & The list of feature selection methods to benchmark. \\
 & \texttt{vary\_param} & Target variable: "features", "instances", or "both". \\
 & \texttt{time\_limit} & Threshold (seconds) to terminate long-running methods. \\ \hline
\end{tabular}
\end{table}

\subsection{Example Usecase}
The following example demonstrates the practical application of the \texttt{FSEVAL} library. In this scenario, we define a custom metric to evaluate structural preservation and execute a comparative benchmark on our feature selection method, namely random baseline, already included in the FSEVAL, and inline definition of variance-based feature selection.

\begin{lstlisting}[language=Python, caption=FSEVAL Integration and Custom Metric Example]
from sdufseval import FSEVAL
import numpy as np
from sklearn.neighbors import NearestNeighbors

def snn_consistency_k5(X_orig, X_sub, y):
    """
    Calculates the average proportion of shared nearest neighbors (k=5) 
    between the original space and the feature-selected subspace.
    """
    k = 5
    k = min(k, X_orig.shape[0] - 1)
    
    def get_nn_indices(data, n_neighbors):
        nbrs = NearestNeighbors(n_neighbors=n_neighbors + 1, 
                                algorithm='auto').fit(data)
        _, indices = nbrs.kneighbors(data)
        return indices[:, 1:]

    nn_orig = get_nn_indices(X_orig, k)
    nn_sub = get_nn_indices(X_sub, k)
    
    intersections = [len(np.intersect1d(nn_orig[i], nn_sub[i])) 
                     for i in range(len(nn_orig))]
    return np.mean(intersections) / k

if __name__ == "__main__":
    DATASETS_TO_RUN = ['colon', 'leukemia', 'prostate_GE']

    # Initialize the evaluator with integrated and custom metrics
    evaluator = FSEVAL(
        output_dir="benchmark_results", 
        avg_steps=5,
        eval_type=["supervised", "unsupervised", "model_agnostic", "custom"],
        custom_metrics={"SNN_K5": snn_consistency_k5}
    )

    # Define methods: custom functions and built-in baselines
    methods_list = [
        {
            'name': 'Random',
            'type': 'unsupervised', 
            'stochastic': True, 
            'func': evaluator.random_baseline
        },
        {
            'name': 'Variance_Baseline',
            'type': 'unsupervised', 
            'stochastic': False, 
            'func': lambda X: np.var(X, axis=0)
        }
    ]
    
    # Execute the integrated evaluation across multiple datasets
    evaluator.run(DATASETS_TO_RUN, methods_list)

    # Perform scalability analysis (Time vs. N and D)
    evaluator.timer(
        methods=methods_list, 
        vary_param='both', 
        time_limit=3600 
    )
\end{lstlisting}

\paragraph{Workflow Explanation}
As shown in the implementation, the workflow is divided into three distinct phases:
\begin{enumerate}
    \item \textbf{Custom Metric Definition:} The \texttt{snn\_consistency\_k5} function is passed to the \texttt{FSEVAL} constructor via the \texttt{custom\_metrics} dictionary. This allows the framework to automatically apply the metric to every feature-selection method during the \texttt{run()} phase.
    \item \textbf{Integrated Benchmarking:} The \texttt{evaluator.run()} method iterates through the specified datasets (\textit{colon, leukemia, prostate\_GE}). It handles the application of both the \texttt{Random} baseline (utilizing the \texttt{avg\_steps=5} to ensure statistical stability) and the \texttt{Variance\_Baseline}.
    \item \textbf{Performance Profiling:} The \texttt{evaluator.timer()} call assesses the computational cost of the methods. By setting \texttt{vary\_param='both'}, the library generates a performance profile showing how execution time scales as either the number of samples or the number of features increase.
\end{enumerate}

\subsection{Interactive Evaluation Dashboard}

\begin{figure}[ht]
    \centering
    \includegraphics[width=\textwidth]{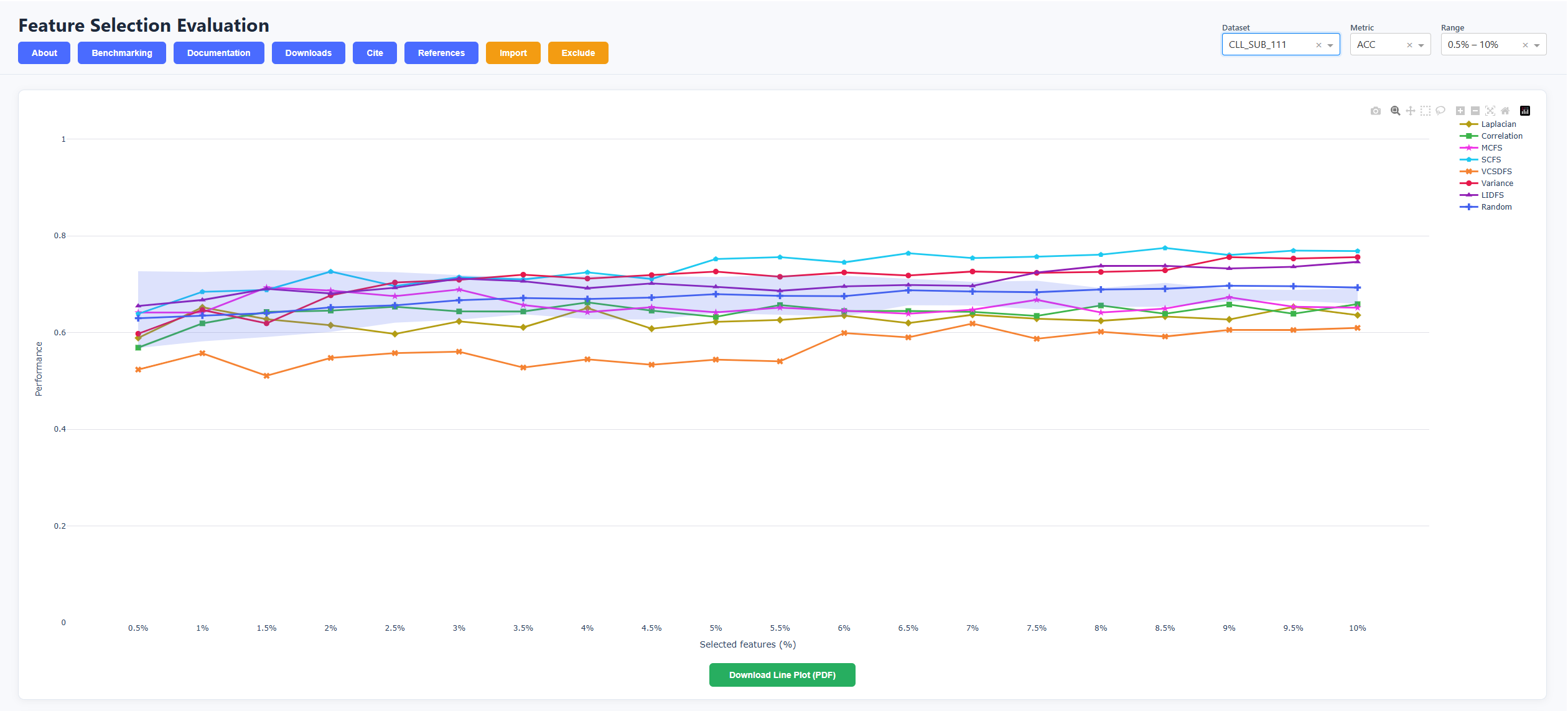}
    \caption{The FSEval Dashboard: An interactive environment for analytical insights and comparisons on feature selection algorithms.}
    \label{fig:dashboard_main}
\end{figure}

The \texttt{FSEval} Dashboard (\url{https://fseval.imada.sdu.dk/}) provides a comprehensive suite of analytic tools for analytical insights and comparisons. As illustrated in Figure~\ref{fig:dashboard_main}, the interface allows for global control over the benchmarking data. High-level options such as \texttt{Import} and \texttt{Exclude} enable users to filter datasets or methods seamlessly across every analysis provided by the platform, or observe the performance of their method against others seamlessly.

\begin{figure}[tb]
    \centering
    \includegraphics[width=\textwidth]{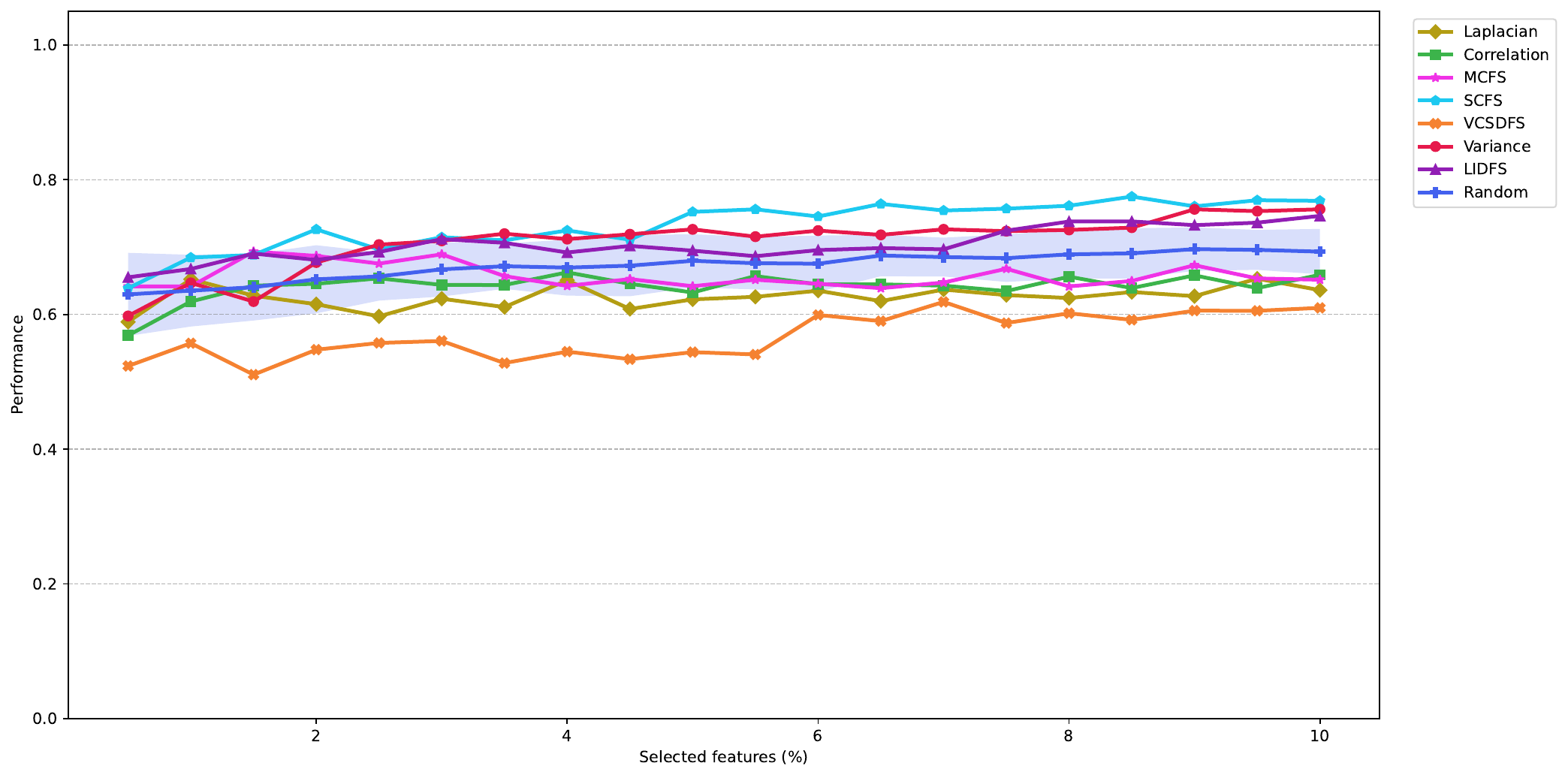}
    \caption{Publication-ready line plot generated by the dashboard, demonstrating the performance trajectory and variance of multiple algorithms.}
    \label{fig:lineplot}
\end{figure}

A critical feature of the dashboard is its ability to generate high-fidelity, publication-ready figures. Figure~\ref{fig:lineplot} showcases a performance profile across a "feature ratio grid," identifying how accuracy (ACC) evolves as the subset size increases from 0.5\% to 10\%. 

\begin{figure}[tb]
    \centering
    % Left Figure (Standard AUC)
    \begin{minipage}[b]{0.45\linewidth}
        \centering
        \includegraphics[width=\linewidth]{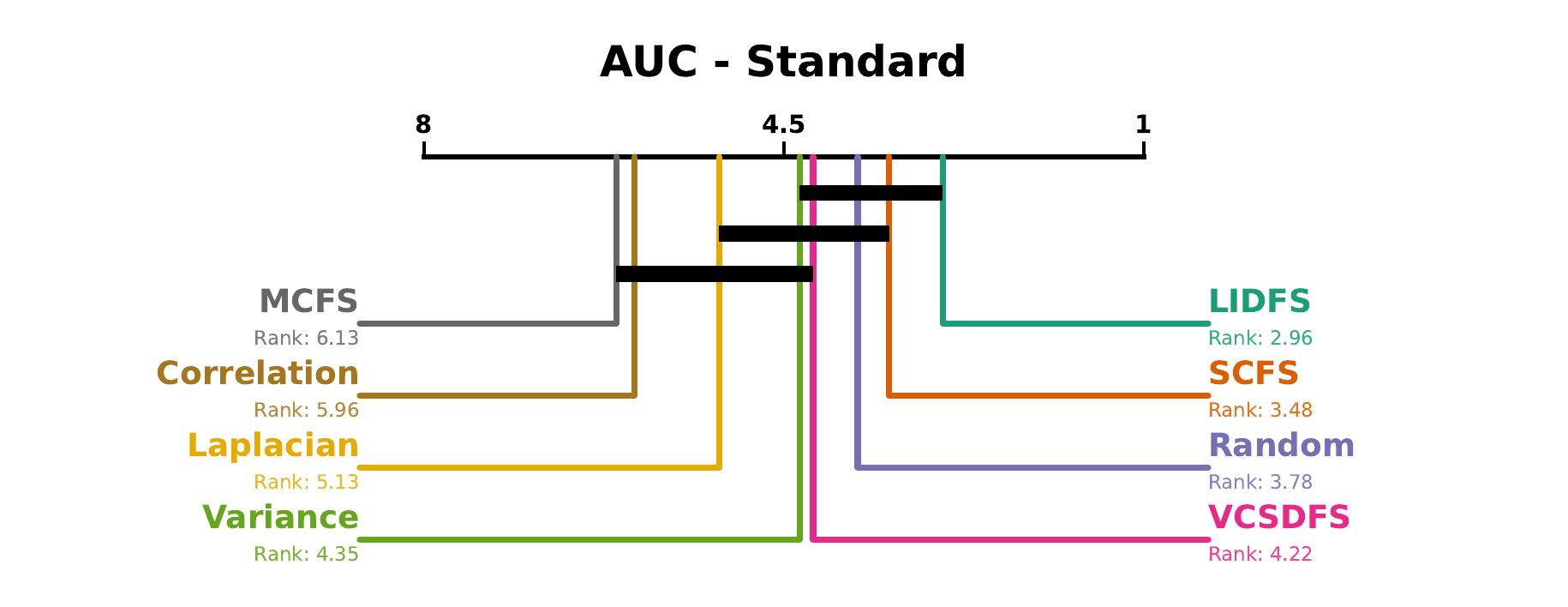}
    \end{minipage}
    \hspace{14pt} % Pushes the two minipages to the left and right edges
    % Right Figure (MARS AUC)
    \begin{minipage}[b]{0.49\linewidth}
        \centering
        \includegraphics[width=\linewidth]{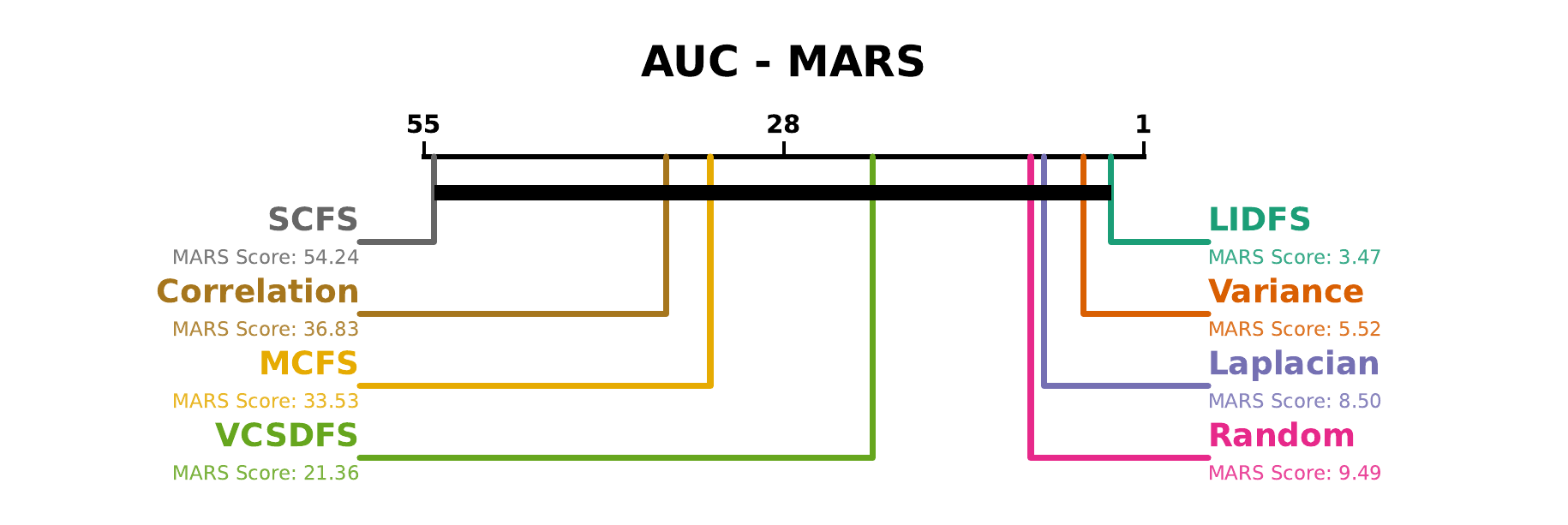}
    \end{minipage}
    
    \caption{Rank Analysis and CD Diagrams provided by \texttt{FSEVAL}.}
    \label{fig:cdd}
\end{figure}

We also provide rank statistics analysis based on the standard rank~\citep{demsar2006statistical} and MARS~\cite{rajabinasab2026marsmagnitudeawarerankstatistics}, as shown in Figure~\ref{fig:cdd}.

\begin{figure}[tb]
    \centering
    \includegraphics[width=\textwidth]{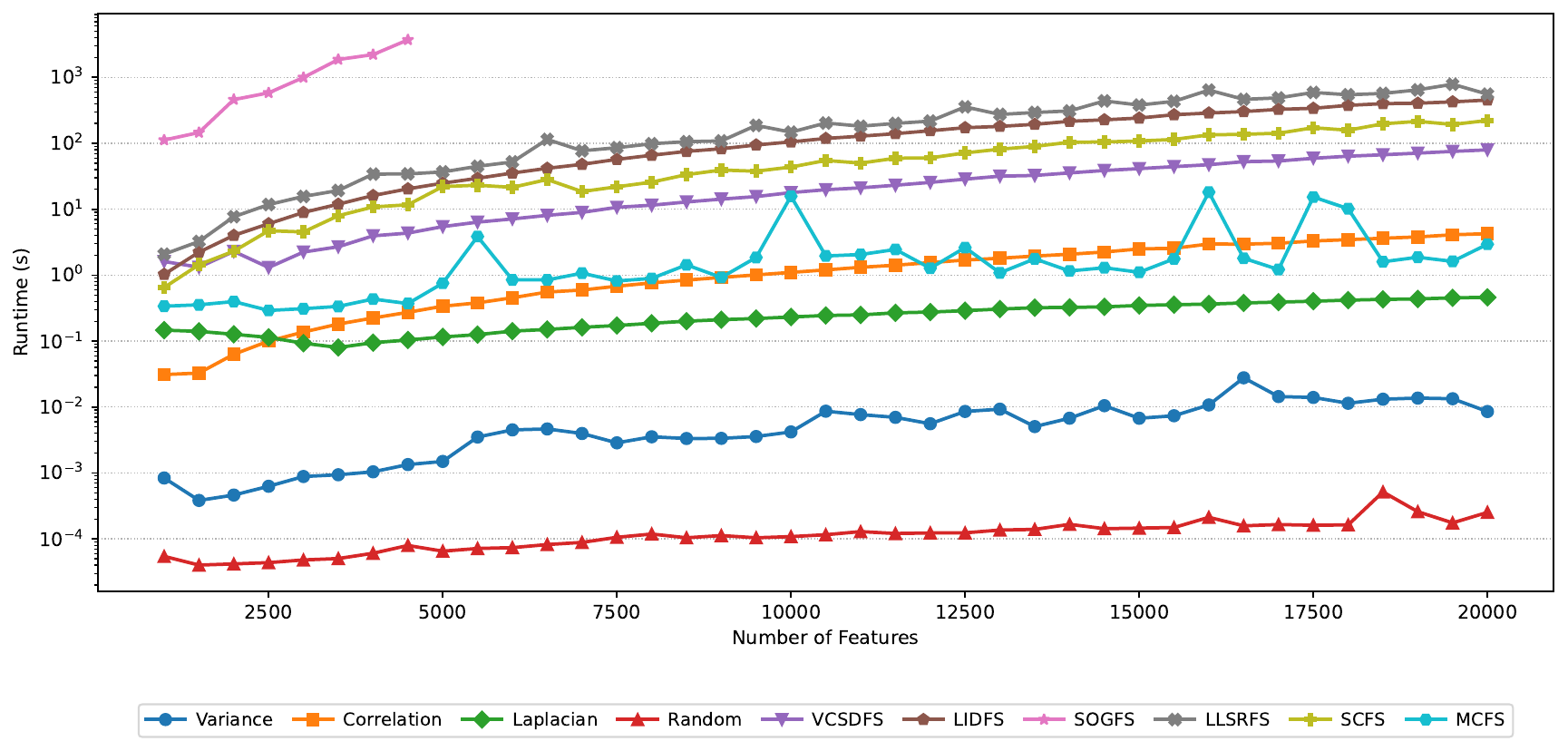}
    \caption{Scalability analysis showing runtime (seconds) against the number of features.}
    \label{fig:runtime_features}
\end{figure}

\begin{figure}[tb]
    \centering
    \includegraphics[width=\textwidth]{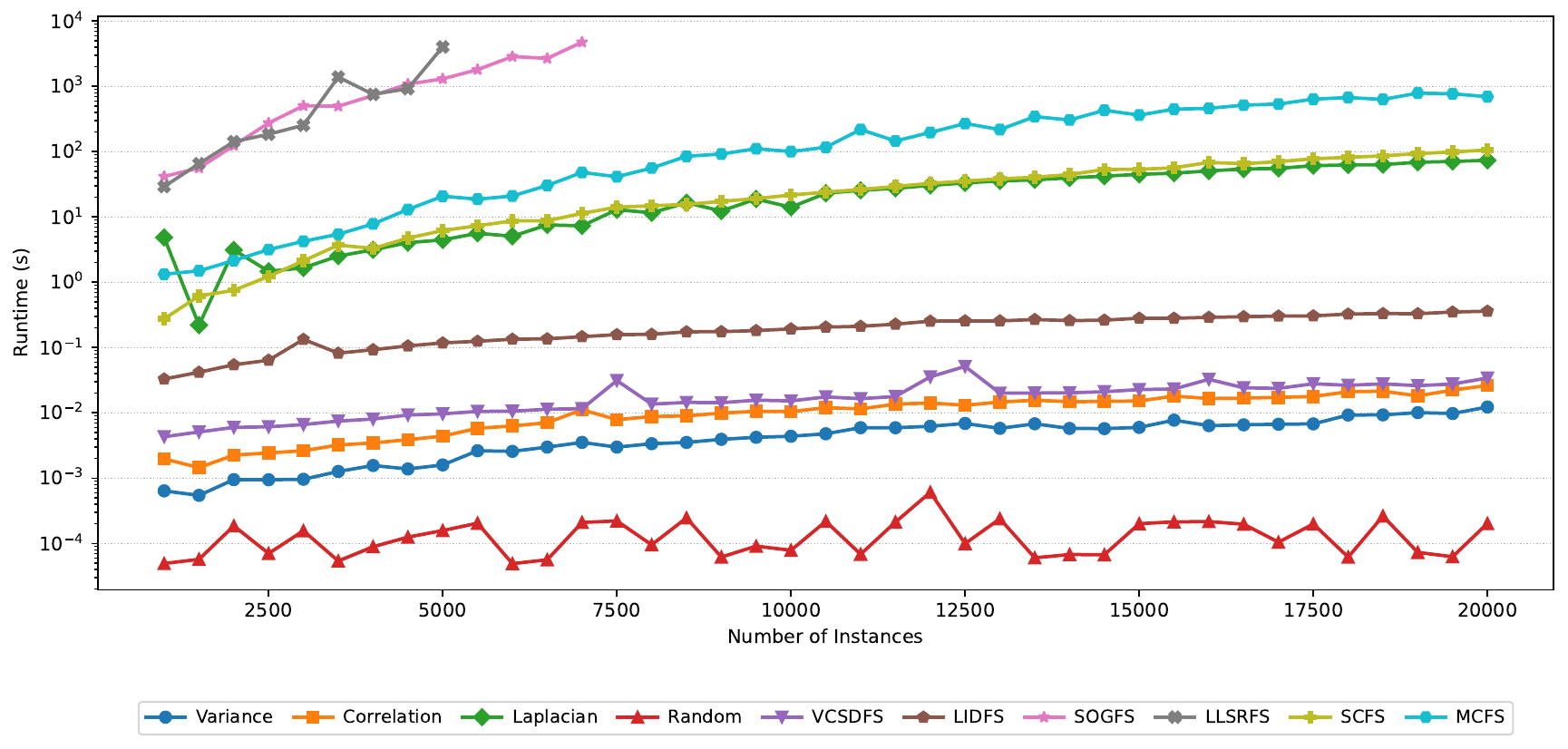}
    \caption{Scalability analysis showing runtime (seconds) against the number of instances.}
    \label{fig:runtime_instances}
    \vspace*{\fill}
\end{figure}

For scalability and runtime analyses, the dashboard visualizes the results of the \texttt{timer} module. Figures~\ref{fig:runtime_features} and \ref{fig:runtime_instances} present these results, contrasting runtime behavior against increasing feature counts and instance sizes. These visualizations are essential for identifying the computational bottlenecks of high-dimensional feature selection methods.

We aim to keep the toolbox and the dashboard updated, keeping it as a nice ecosystem for the evaluation, analysis, visualization, and comparison of the feature selection methods.

\clearpage
\bibliographystyle{plainnat}
\bibliography{sample}

@article{guyon2003introduction,
  author = {Guyon, Isabelle and Elisseeff, André},
  title = {An Introduction to Variable and Feature Selection},
  journal = {Journal of Machine Learning Research},
  volume = {3},
  pages = {1157--1182},
  year = {2003}
}

@article{li2017feature,
  author = {Li, Jundong and Cheng, Kewei and Wang, Suhang and Morstatter, Fred and Trevino, Robert P. and Tang, Jiliang and Liu, Huan},
  title = {Feature Selection: A Data Perspective},
  journal = {ACM Computing Surveys},
  volume = {50},
  number = {6},
  pages = {94:1--94:45},
  year = {2017}
}

@article{mostert2021feature,
  title={A Feature Selection Algorithm Performance Metric for Comparative Analysis},
  author={Mostert, Stephanus J. and Lones, Michael A. and Burger, Andries P.},
  journal={Algorithms},
  volume={14},
  number={3},
  pages={100},
  year={2021},
  publisher={MDPI}
}

@inproceedings{rajabinasab2024fsdem,
  title={A Dynamic Evaluation Metric for Feature Selection},
  author={Rajabinasab, Muhammad and Lautrup, Anton D. and Hyrup, Tobias and Zimek, Arthur},
  booktitle={International Conference on Similarity Search and Applications (SISAP 2024)},
  pages={65--72},
  year={2024},
  organization={Springer}
}

@inproceedings{rajabinasab2025metrics,
  title={Metrics for Inter-Dataset Similarity with Example Applications in Synthetic Data and Feature Selection Evaluation},
  author={Rajabinasab, Muhammad and Lautrup, Anton D. and Zimek, Arthur},
  booktitle={Proceedings of the 2025 SIAM International Conference on Data Mining (SDM 2025)},
  year={2025}
}

@article{nogueira2017stability,
  title={On the Stability of Feature Selection Algorithms},
  author={Nogueira, Sarah and Sechidis, Konstantinos and Brown, Gavin},
  journal={Journal of Machine Learning Research},
  volume={18},
  number={174},
  pages={1--54},
  year={2017}
}

@article{kuncheva2007stability,
  title={A stability index for feature selection},
  author={Kuncheva, Ludmila I.},
  journal={Proceedings of the 25th IASTED International Multi-Conference: Artificial Intelligence and Applications},
  pages={390--395},
  year={2007}
}

@article{pearson1901,
  title={LIII. On lines and planes of closest fit to systems of points in space},
  author={Pearson, Karl},
  journal={The London, Edinburgh, and Dublin Philosophical Magazine and Journal of Science},
  volume={2},
  number={11},
  pages={559--572},
  year={1901},
  publisher={Taylor \& Francis}
}

@inproceedings{bourlard1988,
  title={Auto-association by multilayer perceptrons and singular value decomposition},
  author={Bourlard, Herv{\'e} and Kamp, Yves},
  booktitle={Biological Cybernetics},
  volume={59},
  pages={291--294},
  year={1988}
}

@article{vandermaaten2008,
  title={Visualizing data using t-SNE},
  author={Van der Maaten, Laurens and Hinton, Geoffrey},
  journal={Journal of Machine Learning Research},
  volume={9},
  number={11},
  pages={2579--2605},
  year={2008}
}

@article{mcinnes2018,
  title={UMAP: Uniform Manifold Approximation and Projection for Dimension Reduction},
  author={McInnes, Leland and Healy, John and Melville, James},
  journal={arXiv preprint arXiv:1802.03426},
  year={2018}
}

@article{dy2003feature,
  title={Feature selection for unsupervised learning},
  author={Dy, Jennifer G. and Brodley, Carla E.},
  journal={Journal of Machine Learning Research},
  volume={5},
  pages={845--889},
  year={2004}
}

@article{lloyd1982,
  title={Least squares quantization in PCM},
  author={Lloyd, Stuart},
  journal={IEEE transactions on information theory},
  volume={28},
  number={2},
  pages={129--137},
  year={1982},
  publisher={IEEE}
}

@article{demsar2006statistical,
  title={Statistical comparisons of classifiers over multiple data sets},
  author={Dem{\v{s}}ar, Janez},
  journal={Journal of Machine Learning Research},
  volume={7},
  number={1},
  pages={1--30},
  year={2006}
}

@article{REIS201710,
title = {featsel: A framework for benchmarking of feature selection algorithms and cost functions},
journal = {SoftwareX},
volume = {6},
pages = {10-15},
year = {2017},
issn = {2352-7110},
doi = {https://doi.org/10.1016/j.softx.2017.01.002},
Xurl = {https://www.sciencedirect.com/science/article/pii/S2352711017300286},
author = {Benedito José {Vieira Reis} and Ricardo {Pinto Ferreira} and {Chun-Wei} Tsai and {Yung-Che} Tseng and Ivan {Izonin} and Aras {Asaad} and Shadi {Alawneh}}
}

@article{zou2026feature,
  title={Feature selection based on rough diversity entropy},
  author={Zou, Xiongtao and Dai, Jianhua},
  journal={Pattern Recognition},
  volume={170},
  pages={112032},
  year={2026},
  publisher={Elsevier}
}

@article{dai2026online,
  title={Online multi-label streaming feature selection by affinity significance, affinity relevance and affinity redundancy},
  author={Dai, Jianhua and Xu, Duo and Zhang, Chucai},
  journal={Pattern Recognition},
  volume={170},
  pages={111990},
  year={2026},
  publisher={Elsevier}
}

@article{jain2026enfestdroid,
  title={EnFeSTDroid: Ensembled feature selection techniques based Android malware detection},
  author={Jain, Suruchi and Goyal, Hemant and Arora, Anshul and Kumar, Dhirendra},
  journal={Computers and Electrical Engineering},
  volume={129},
  pages={110763},
  year={2026},
  publisher={Elsevier}
}

@article{rajabinasab2026marsmagnitudeawarerankstatistics,
  author       = {Muhammad Rajabinasab and
                  Afsaneh M. Nejad and
                  Arthur Zimek},
  title        = {{MARS:} Magnitude-Aware Rank Statistics},
  journal      = {CoRR},
  volume       = {abs/2605.23563},
  year         = {2026} 
}

@article{rajabinasab2026worserandomimportancebaseline,
  author       = {Muhammad Rajabinasab and
                  Michael E. Houle and
                  Oussama Chelly and
                  Arthur Zimek},
  title        = {Worse than Random: The Importance of a Baseline for Unsupervised Feature
                  Selection},
  journal      = {CoRR},
  volume       = {abs/2605.22973},
  year         = {2026} 
}

\end{document}